\documentclass{article}
\usepackage{spconf,amsmath,graphicx}
\usepackage{amssymb,amsfonts}
\usepackage{algorithmic}
\usepackage{multirow}
\usepackage{adjustbox}
\usepackage{comment}

\title{Latent graph encoding of multimodal neuroimaging features with generative AI architectures }
\name{Ishaan Batta$^{*+}$, Meenu Ajith$^*$, Vince Calhoun \thanks{* these authors contributed equally to this work\\ $+$ Corresponding author: Ishaan Batta (ibatta@gsu.edu); \\ This work was supported by NIH grants 1R01AG090597, R01AG073949 awarded to V. Calhoun}}
\address{Center for Translational Research in Neuroimaging and Data Science (TReNDS): \\ Georgia State University, Georgia Institute of Technology, and Emory University, Atlanta, USA}
\begin{document}
%
\maketitle
\begin{abstract}
While generative models enable encoding of complex neuroimaging data for feature generation and reconstruction, developing optimal architectural frameworks with appropriate encoding and latent space processes is crucial for studying structural and functional properties of the brain. We design a multimodal generative framework for structural and functional magnetic resonance imaging (MRI) features through systematic evaluation of encoding strategies, latent multimodal fusion, and generative model selection. Using structural gray matter volume (GMV) and static functional network connectivity (sFNC) features from a large neuroimaging dataset, we analyze generative frameworks involving variational autoencoders (VAEs), transformers, generative adversarial networks (GANs), and diffusion models. Architectures that employ modality-aware graph encoding of functional connectivity into a lower-dimensional latent space outperform vectorized encoders or direct data space approaches. The proposed multimodal graph VAE (gMMVAE) surpasses alternative generative variants across multiple metrics for generation fidelity, reconstruction quality, efficiency, and latent space discriminability, highlighting its potential for robust multimodal neuroimaging analysis.
\end{abstract}
\begin{keywords}
Generative AI, Multimodal learning, Neuroimaging, Graph Neural Networks, Data Reconstruction 
\end{keywords}
\section{Introduction}
\label{sec:intro}
\vspace{-6pt}
Understanding how brain structure and function relate to externally observed clinical and demographic variables has driven the development of neuroimaging studies leveraging magnetic resonance imaging (MRI) and modern machine learning methods \cite{nenning2022machine, davatzikos2019machine}. While one major line of studies focus on learning the associations for accurately predicting target variables \cite{nenning2022machine}, many unsupervised and semi-supervised learning frameworks have been developed to create low-dimensional representations for further associative analysis \cite{du2020neuromark, mwangi2014review}. Deep learning and generative learning models have enabled a more sophisticated way to map complex neuroimaging features into low-dimensional latent encodings for feature-to-target prediction \cite{avbervsek2022deep} or feature-to-feature mappings for reconstruction and generation \cite{gong2023generative}. Information encoded in the latent space of these models can be utilized and studied for extracting meaningful associative patterns related to assessment variables and the features being encoded \cite{asperti2023comparing}. This can be achieved by studying the latent space itself, or also the prediction and reconstruction outputs yielded by the model.

Functional MRI (fMRI) models brain dynamics and the connectivity between different brain regions \cite{fornito2015connectomics}, while structural MRI (sMRI) captures static features such as gray matter volume (GMV) \cite{calhoun2016multimodal}. Multimodal learning integrating both modalities provides more comprehensive views of brain dynamics \cite{calhoun2016multimodal} compared to unimodal analysis. Deep generative models encode these features into latent spaces either through direct voxel-level processing \cite{avbervsek2022deep, gong2023generative}, or using pre-computed features summarized at the level of brain regions or components, such as static functional network connectivity (sFNC) matrices and GMV \cite{du2020neuromark}. The choice of encoders is crucial for downstream learning tasks as it determines the performance and explainability of the model. In case of sFNC features, graph encoders offer a relevant encoding method for functional connectivity \cite{bessadok2022graph}, and these features enable tasks involving prediction, reconstruction, and generation. Recently, generative models like variational autoencoders (VAEs) \cite{kim2021representation}, generative adversarial networks (GANs) \cite{wang2023applications}, transformers \cite{zhu2025transformer}, and latent diffusion models \cite{pinaya2022brain} offer a promising way for generating and reconstructing neuroimaging features while utilizing lower-dimensional latent encodings for better performance \cite{gong2023generative}.

This paper introduces a unified framework for multimodal generative modeling of structural and functional neuroimaging data. The main contributions of this work are threefold: (i) we propose a modality aware multimodal generative framework that utilizes the graph structure of functional connectivity while jointly modeling structural and functional MRI features; (ii) we conduct a systematic evaluation of graph based generative architectures against vectorized and data space baselines, providing a comprehensive comparison of encoding strategies, latent fusion and generation; and (iii) we introduce a condition guided multimodal graph VAE (gMMVAE) that integrates subject-level covariates and achieves consistent performance by preserving variance patterns in generated features while delivering improved efficiency and generation quality compared to alternative generative models.

\begin{figure*}[htbp]
    \centering
    \includegraphics[width=\linewidth]{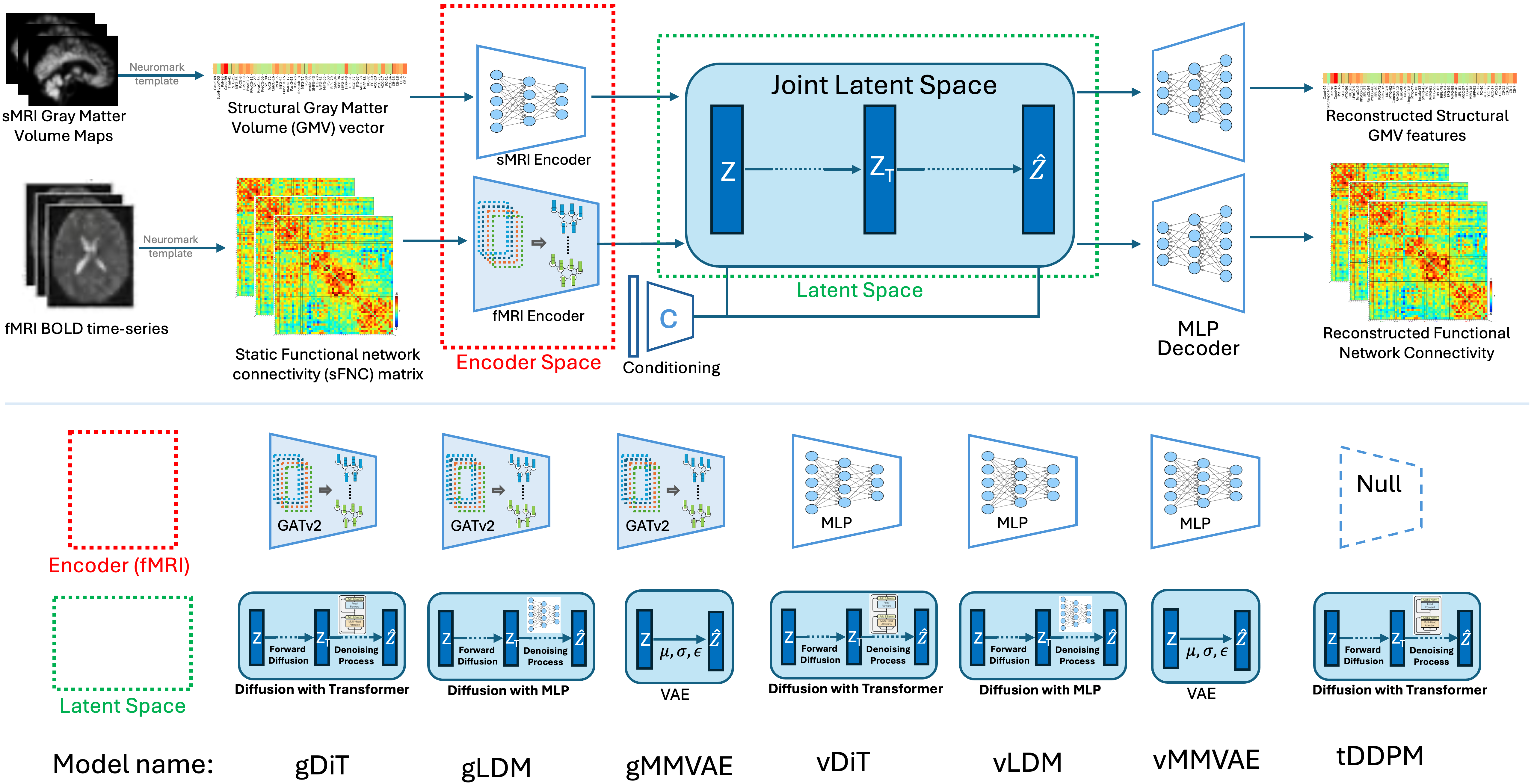}
    \caption{Overview of the latent generative models architectures based on the choice of encoders and latent space processes.}
    \label{fig:main}
\end{figure*}

\vspace{-6pt}
\section{Methods}
\label{sec:format}
\vspace{-6pt}
We present a framework for multimodal generative modeling and fusion of neuroimaging data, where each subject is characterized by both structural and functional MRI information. The fMRI modality is represented using sFNC matrices, while the sMRI modality is represented as a vector of GMV features. Let $(\mathbf{X}^f, \mathbf{X}^s )$ denote paired samples where $\mathbf{X}^f \in \mathbb{R}^{53 \times 53}$ and $\mathbf{X}^s \in \mathbb{R}^{53}$. The primary objective is to learn a joint generative model capable of reconstructing and generating both modalities while preserving modality-specific structure and coherence. 
\vspace{-6pt}
\subsection{Modality-aware encoders and multimodal fusion}
Each modality is processed using an encoder designed for its specific underlying structure. For fMRI, the sFNC matrix is modeled as a weighted graph, where nodes correspond to brain regions and edge weights reflect connectivity strengths. A multilayer Graph Attention Network (GATv2) \cite{brody2021attentive} is employed to effectively capture higher-order dependencies and preserve network topology. The GATv2 encoder adaptively aggregates information from neighboring nodes through attention-based message passing, resulting in a latent representation that captures the relational structure of the functional brain networks. In contrast, the sMRI GMV features are represented as fixed-length vectors without any relational structure and are therefore encoded using a multi-layer perceptron (MLP). This encoder efficiently captures the global structural variations across the brain regions.

Modality-specific approximate posterior distributions, denoted as $q_{\phi_f}(\mathbf{z}_f|\mathbf{X}^f)$ and $q_{\phi_s}(\mathbf{z}_s|\mathbf{X}^s)$, are produced by each encoder, and are parameterized by a modality-specific mean and variance. The multimodal fusion is implemented using a multimodal variational autoencoder (MMVAE) \cite{shi2019variational} that integrates the modality-specific latent distributions into a shared latent space. The Mixture-of-Experts (MoE) mechanism in MMVAE combines modality-specific posteriors as follows:
\begin{equation}
    q_{\phi}(\mathbf{z}|\mathbf{X}^f,\mathbf{X}^s)=\sum_{m \in {f,s}} \alpha_m q_{\phi_m}(\mathbf{z}_m|\mathbf{X}^m)
\end{equation}
where $\alpha_m$ is the learned mixture coefficients. The MMVAE is trained by maximizing the evidence lower bound (ELBO):
\begin{equation}
\begin{aligned}
\mathcal{L}_{\text{MMVAE}} =
&\mathbb{E}_{q_\phi(\mathbf{z}|\mathbf{X}^f,\mathbf{X}^s)}
\Big[
\log p_{\theta_f}(\mathbf{X}^f|\mathbf{z})
+ \log p_{\theta_s}(\mathbf{X}^s|\mathbf{z})
\Big] \\
&- \beta \, D_{KL}\big(
q_\phi(\mathbf{z}|\cdot) \,\|\, p(\mathbf{z})
\big)
\end{aligned}
\end{equation}
where $p(\mathbf{z})$ is a standard Gaussian prior and $\beta$ controls the trade-off between reconstruction fidelity and latent regularization. Finally, the MLP-based decoders are employed to reconstruct each modality from the shared latent variable $z$. 

\subsection{Condition-guided inference and generation}
The proposed MMVAE explicitly incorporates conditioning variables at both encoder and decoder stages in addition to latent multimodal fusion. These conditioning variables, denoted by $\mathbf{c}$, correspond to subject-level metadata. In this study, sex is used as a representative covariate due to its biological relevance. However, the framework can be extended to incorporate additional variables such as age and clinical measures. At the encoder level, conditioning is directly integrated into the approximate posterior distributions. For each modality $m \in \{f, s\}$, the encoder produces an intermediate embedding $h_{m}$, which is combined with a learned embedding of the conditioning vector $\mathbf{c}_m$. The mean and variance of the modality-specific posteriors are defined as follows:
\begin{equation}
\boldsymbol{\mu}_m = f_{\mu}(h_m) + g(\mathbf{c}_m),
\qquad
\log \boldsymbol{\sigma}_m^2 = f_{\sigma}(h_m) + g(\mathbf{c}_m),
\end{equation}
Where $g(.)$ denotes a conditioning network consisting of fully connected layers. Through this additive conditioning, the latent distributions are guided by subject-specific covariates while preserving the representational capacity of the modality encoders. Subsequently, during the decoder stage, conditioning is incorporated using feature-wise linear modulation (FiLM) \cite{perez2018film}. The latent activations are modulated for each decoder layer as
\begin{equation}
\operatorname{FiLM}(h, \mathbf{c})= \gamma(\mathbf{c}) \odot h + \beta(\mathbf{c}),
\end{equation}
where $\gamma(\cdot)$ and $\beta(\cdot)$ are learned linear projections of the conditioning vector. This conditioning mechanism allows fine-grained control of the generative process, enabling decoders to adjust reconstruction dynamics based on subject-level attributes. Therefore, this dual conditioning strategy enables controlled and expressive multimodal generation without increasing the latent dimensionality. 

A comparative analysis was conducted using multiple learning models for reconstruction and generation tasks. We utilized baseline generative models that operate directly in the data space, including denoising diffusion probabilistic models (DDPMs) with a transformer backbone and Wasserstein GAN with gradient penalty (WGAN-GP). Latent space models include multimodal variational autoencoders (MMVAE), latent diffusion models (LDM), and diffusion transformers (DiT). A prefix indicating the encoder backbone was used for each model: “v” for MLP-based (vMMVAE, vLDM, vDiT), “t” for transformer-based (tDDPM), and “g” for graph-based (gMMVAE, gLDM, gDiT) architectures. Architectural configurations, training schedules, and hyperparameters of the baseline models are detailed in Table.~\ref{fig:hyper}. The overall network architecture is shown in Fig.~\ref{fig:main}.
\vspace{-12pt}
\begin{table}[ht]
\centering
\caption{Hyperparameter configuration for baseline models}
\label{fig:hyper}
\begin{adjustbox}{max width=\columnwidth}
\begin{tabular}{|l|c|c|c|c|c|}
\hline
\textbf{Hyperparameter}                                                                  & \textbf{tDDPM} & \textbf{WGAN-GP}                                                           & \textbf{vDiT}                                              & \textbf{vLDM} & \textbf{vMMVAE}                                               \\ \hline
\textbf{Architecture}                                                                    & Transformer    & \begin{tabular}[c]{@{}c@{}}Generator- MLP\\ Discriminator-MLP\end{tabular} & \begin{tabular}[c]{@{}c@{}}Transformer+\\ MLP\end{tabular} & MLP           & MLP                                                           \\ \hline
\textbf{Hidden dimension}                                                                & 512            & 256/512                                                                    & 128                                                        & 128           & 256                                                           \\ \hline
\textbf{No. of layers}                                                                   & 6              & 4/5                                                                        & 4                                                          & 4             & \begin{tabular}[c]{@{}c@{}}Encoder-5\\ Decoder-2\end{tabular} \\ \hline
\textbf{Attention heads}                                                                 & 8              & -                                                                          & 4                                                          & -             & -                                                             \\ \hline
\textbf{Latent dimension}                                                                & -              & 100                                                                        & 40                                                         & 40            & 40                                                            \\ \hline
\textbf{Beta schedule}                                                                   & Quadratic      & -                                                                          & Quadratic                                                  & Quadratic     & -                                                             \\ \hline
\textbf{Timesteps}                                                                       & 1000           & -                                                                          & 100                                                        & 100           & -                                                             \\ \hline
\textbf{Optimizer}                                                                       & AdamW          & AdamW                                                                      & AdamW                                                      & AdamW         & AdamW                                                         \\ \hline
\textbf{Learning rate}                                                                   & 1e-4           & 1e-4/5e-5                                                                  & 1e-4                                                       & 1e-4          & 1e-4                                                          \\ \hline
\textbf{Weight decay}                                                                    & 1e-4           & 0                                                                          & 1e-5                                                       & 1e-5          & 1e-5                                                          \\ \hline
\textbf{Epochs}                                                                          & 1000           & 500                                                                        & 200                                                        & 200           & 100                                                           \\ \hline
\textbf{\begin{tabular}[c]{@{}l@{}}Gradient penalty/\\ Critical iterations\end{tabular}} & -              & 10/5                                                                       & -                                                          & -             & -                                                             \\ \hline
\end{tabular}
\end{adjustbox}
\end{table}

\section{Experiments}
\vspace{-6pt}
\label{sec:pagestyle}
\subsection{Datasets and Evaluation Metrics}
\label{sec:data}
\vspace{-6pt}
This study used the fMRI and sMRI data from 10, 000 UK Biobank subjects (5376 males and 4624 females) \cite{miller2016multimodal}. Preprocessed rs-fMRI data was passed through a fully automated spatially constrained ICA pipeline known as NeuroMark \cite{du2020neuromark} to obtain subject-specific spatial maps and time courses (TCs). The sFNC features were computed as Pearson correlations between TCs of extracted brain components. We used the Neuromark\_fMRI\_1.0 template, consisting of 53 data-driven brain components divided into 7 functional subdomains: Subcortical (SC), Auditory (AUD), Sensorimotor (SM), Visual (VIS), Cognitive Control (CC), Default Mode Network (DMN), and Cerebellar (CB). For sMRI features, the mean gray matter volume (GMV) values were extracted within the 53 NeuroMark brain component masks. Moreover, although we use the 53-component NeuroMark template, the proposed graph-based formulation is not tied to a specific parcellation and can be directly applied to other atlases with different numbers of regions by adapting the input dimensionality.

Model performance was evaluated across 3 aspects: (1) reconstruction fidelity using Mean Squared Error (MSE), Frobenius norm, Pearson correlation, Structural Similarity Index (SSIM), and Peak Signal-to-Noise Ratio (PSNR); (2) generation quality using Maximum Mean Discrepancy (MMD), Wasserstein Distance (WD), and Kullback-Leibler (KL) divergence to measure distributional alignment between generated and real samples; and (3) latent space discriminability using random forest classifier on the joint latent embeddings with sex labels, yielding accuracy, precision, recall, and F1 score from 5 fold cross validation. 

\vspace{-12pt}
\subsection{Experimental setup}
\label{sec:set}
\vspace{-6pt}
The experiments were conducted using a selection of generative models, including tDDPM, WGAN-GP, DiT, LDM, and MMVAE, focusing on graph variants (gDiT, gLDM, and gMMVAE) that utilize the topological structure of brain connectivity networks for multimodal fusion. Graph-based models received $53\times53$ sFNC matrices from the fMRI modality and $53\times1$ vectors of GMV from the sMRI modality as input. The fMRI encoder used GATv2 with 5 layers for sFNC matrices, while the sMRI encoder employed a 5-layer MLP for GMV vectors, both with 256 hidden dimensions. The modality encodings were fused via MoE into a 40-dimensional latent space, and they were decoded by modality-specific 2-layer MLPs with 256 hidden dimensions. The gMMVAE architecture was trained using the AdamW optimizer with a learning rate of $2 \times 10^{-4}$ and weight decay of $1 \times 10 ^{-5}$ for 100 epochs. Meanwhile, gLDM and gDiT used two stage training consisting of MMVAE pretraining followed by diffusion refinement. The gLDM denoiser was a 4-layer MLP with hidden and conditioning dimensions of 128, whereas gDiT used a 4-layer transformer with 4 attention heads, also featuring hidden and conditioning dimensions of 128. Both models were trained for 200 epochs with AdamW optimizer using a learning rate of $1\times 10^{-4}$ and a weight decay of $1 \times 10^{-5}$. These models also used a quadratic beta schedule with 100 timesteps with $\beta$ increasing from $1 \times10^{-4}$  to $1.5\times10^{-3}$. All three models were trained on paired fMRI-sMRI samples using a batch size of 16, with the data split into 80\% for training, 10\% for validation, and 10\% for testing. NVIDIA V100 GPUs were used with Optuna hyperparameter optimization.


\vspace{-12pt}
\section{Results}
\vspace{-6pt}
\label{sec:result}
\subsection{Reconstruction quality assessment}
\begin{figure}[htbp]
\centering

\setlength\tabcolsep{0pt}
\begin{tabular}{p{.2cm} c c c c}
 & Real Data & MMVAE & LDM & DiT \\

\rotatebox[origin=c]{90}{\small{mean}}    
& \raisebox{-0.5\totalheight}
 {\includegraphics[width=0.23\linewidth]{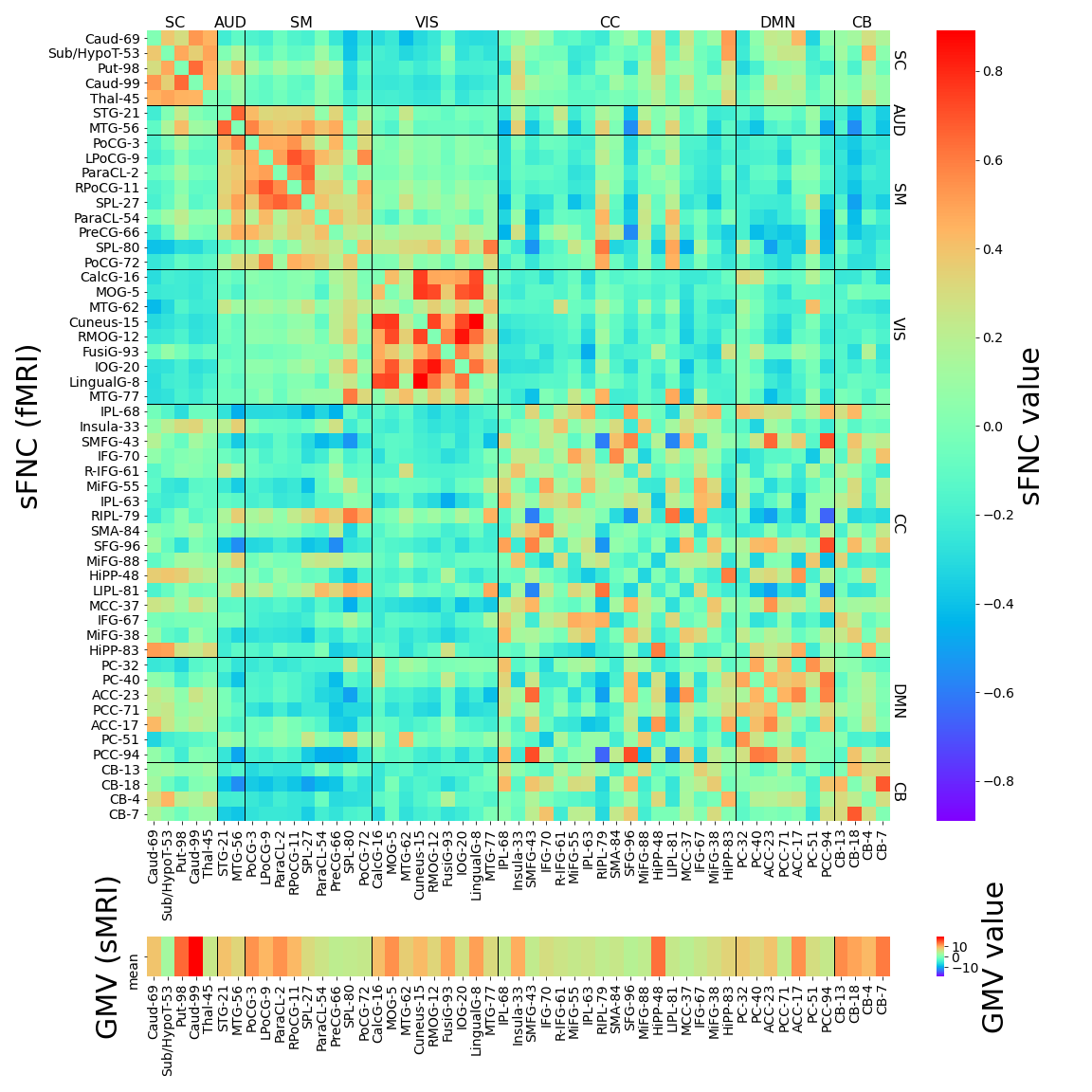}}
& \raisebox{-0.5\totalheight}
 {\includegraphics[width=0.23\linewidth]{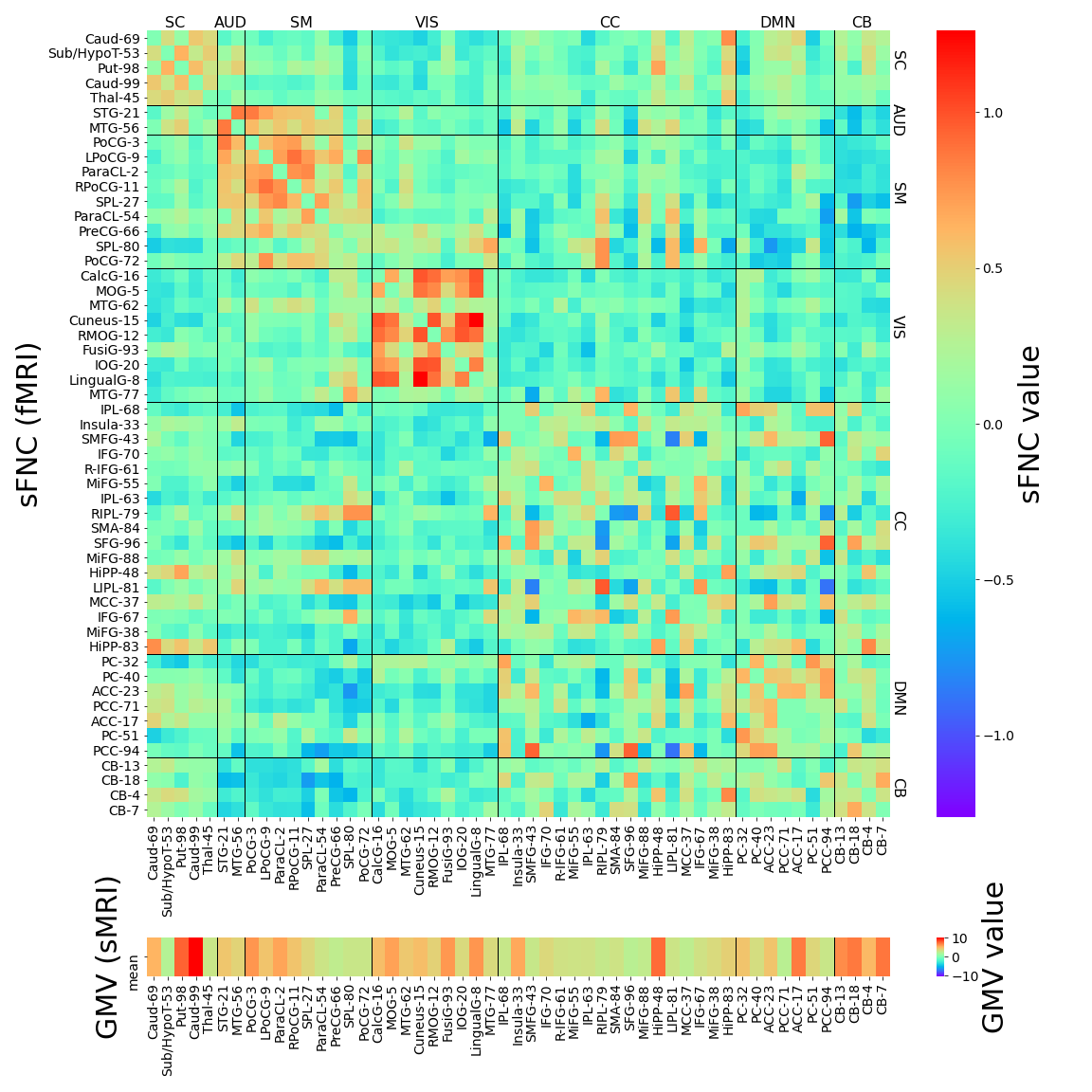}}
& \raisebox{-0.5\totalheight}
 {\includegraphics[width=0.23\linewidth]{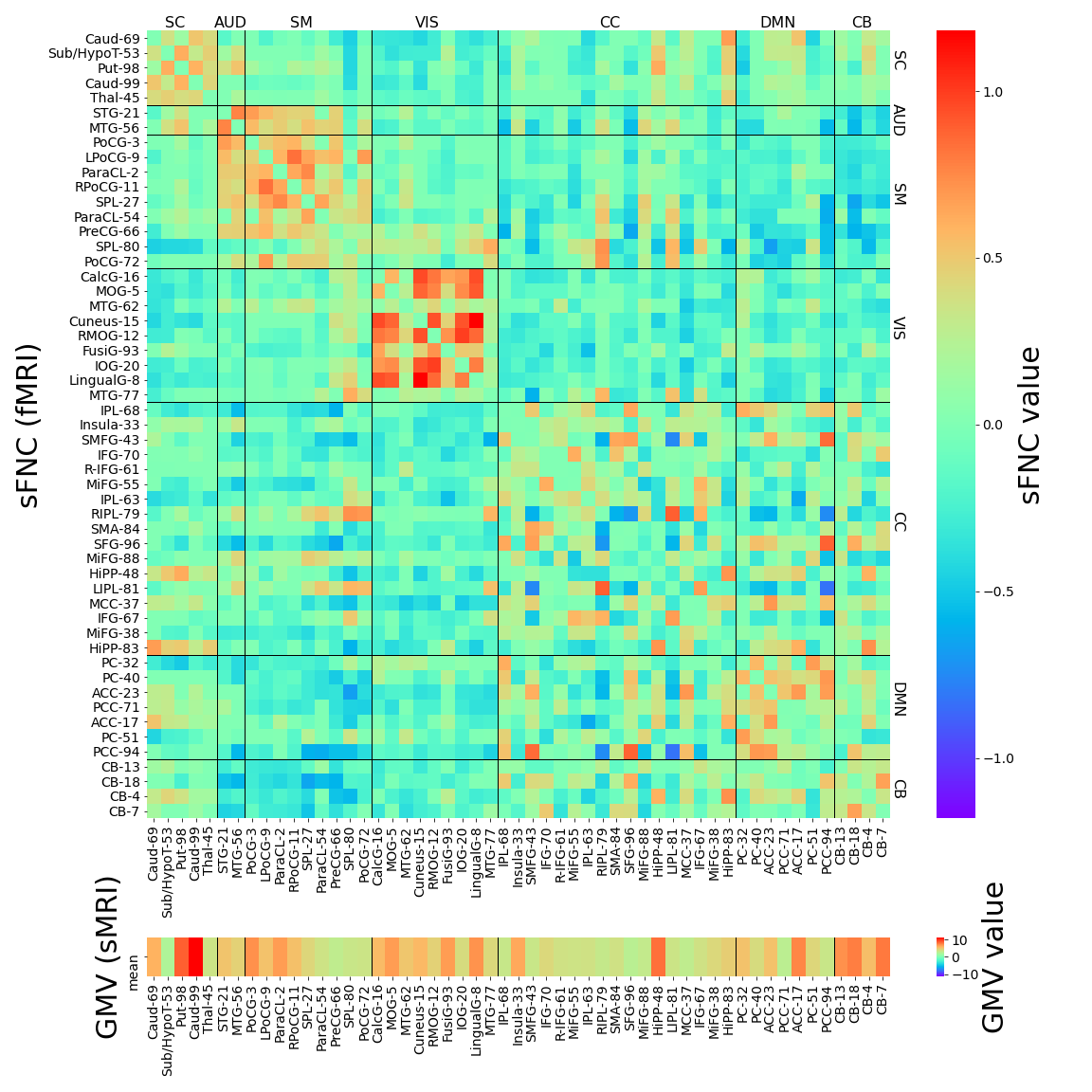}}
& \raisebox{-0.5\totalheight}
 {\includegraphics[width=0.23\linewidth]{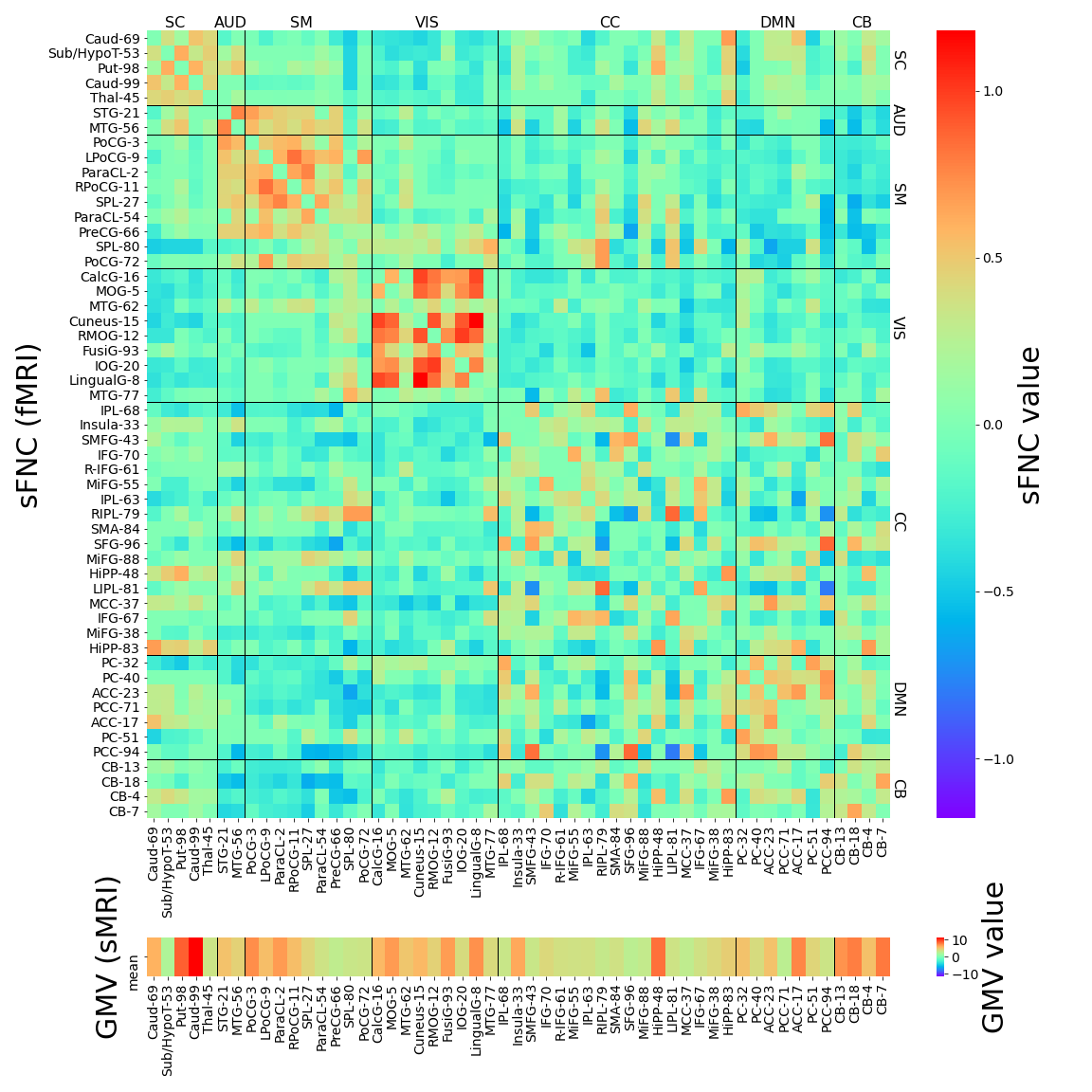}} \\

\rotatebox[origin=c]{90}{\small{std}}    
& \raisebox{-0.5\totalheight}
 {\includegraphics[width=0.23\linewidth]{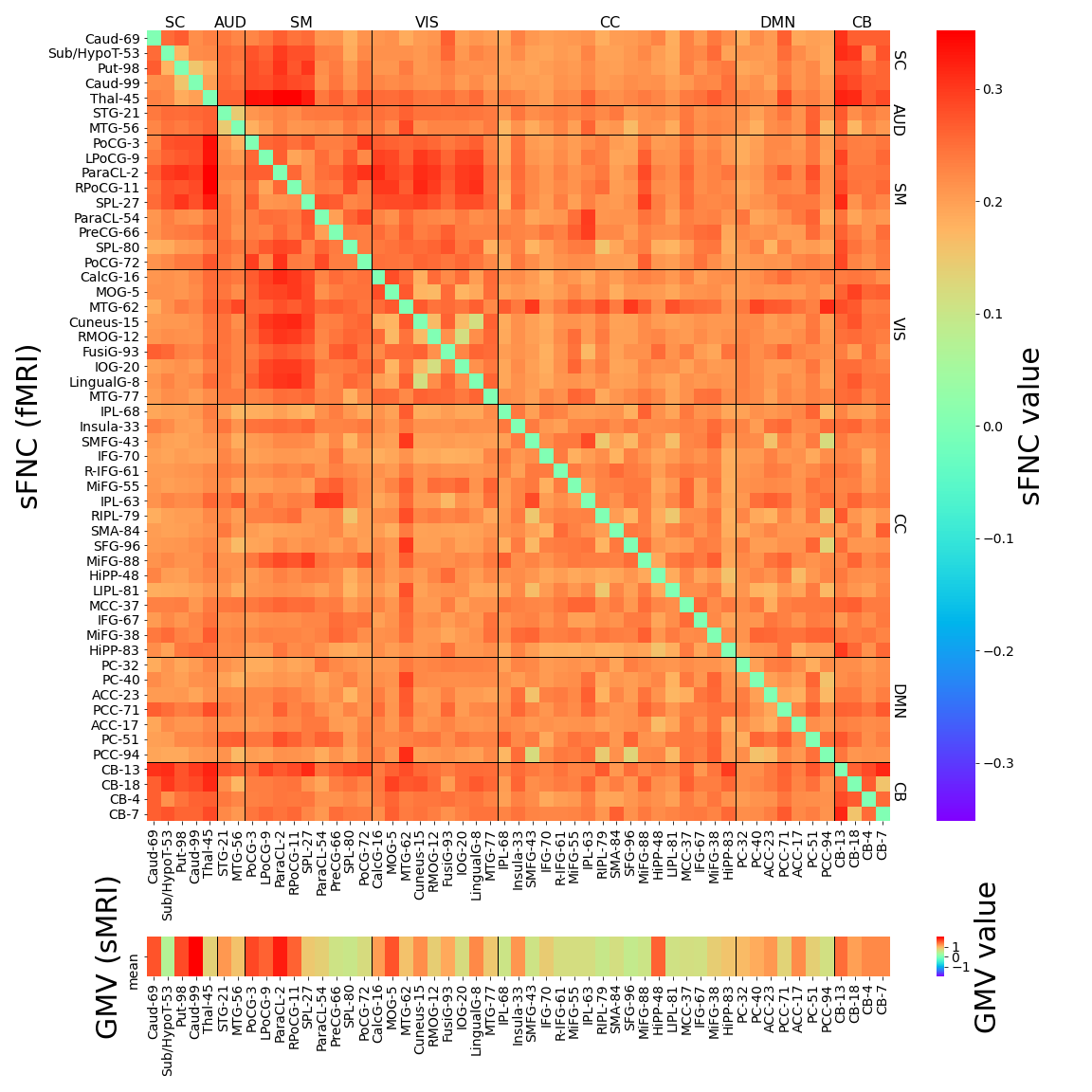}}
& \raisebox{-0.5\totalheight}
 {\includegraphics[width=0.23\linewidth]{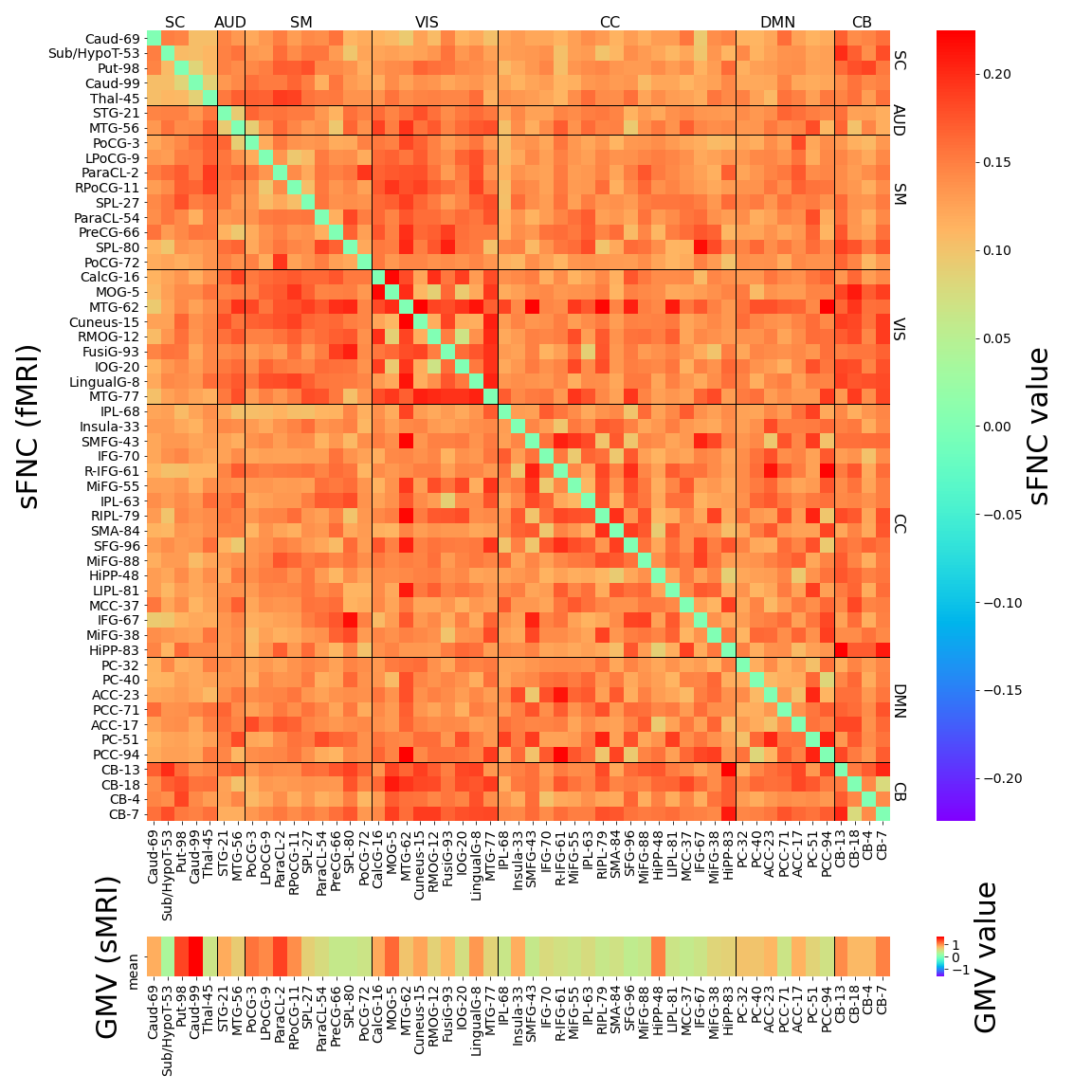}}
& \raisebox{-0.5\totalheight}
 {\includegraphics[width=0.23\linewidth]{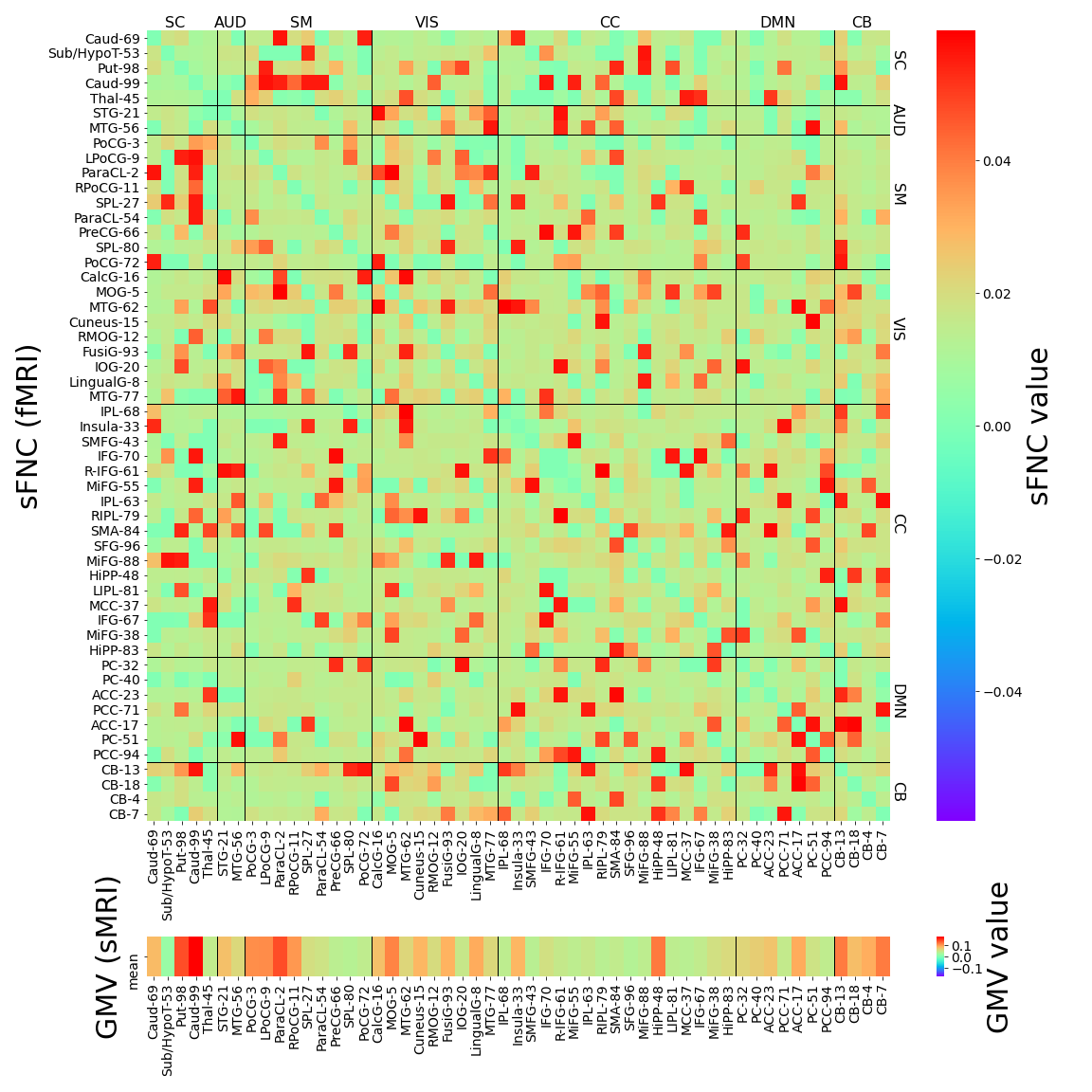}}
& \raisebox{-0.5\totalheight}
 {\includegraphics[width=0.23\linewidth]{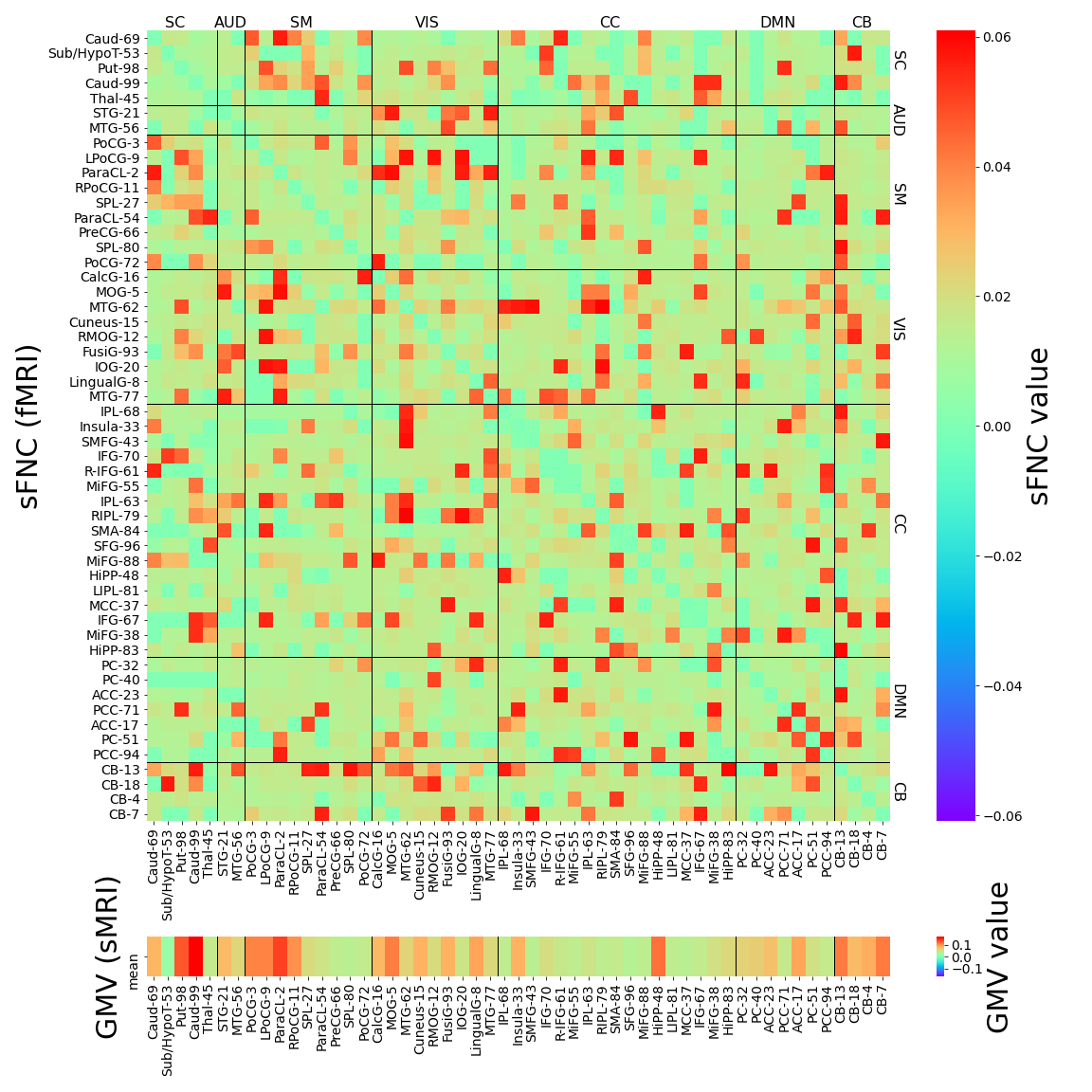}} \\

\end{tabular}

\caption{Mean and std values of sFNC and GMV features from real and generated data using graph-based multimodal VAE (gMMVAE), latent diffusion model (gLDM), and diffusion transformer (gDiT). It can be noted that gMMVAE is better at preserving both the mean and standard deviation in the data.}
\label{fig:overview}
\end{figure}

\vspace{-6pt}
A comprehensive evaluation of reconstruction quality was conducted across 8 generative approaches in Table.~\ref{fig:recon}, showing that the graph-based models substantially outperform their MLP-based and baseline counterparts. For fMRI, gMMVAE, gLDM, and gDiT achieve the lowest MSE, superior Frobenius norms, highest correlations, and SSIM values double those of the standard DDPM variant. Graph-based models reach 20.7dB PSNR versus 14.5-16.19 dB for GANs and tDDPM, confirming strong preservation of functional connectivity patterns. For sMRI reconstruction, all MMVAE-based models exhibited strong performance with high correlations $0.99 \pm 0.01$, low MSE, and PSNR above 30 dB. indicating an improvement compared to the DDPM variant. Comparably, WGAN-GP achieves competitive sMRI reconstruction with a correlation of $0.97 \pm 0.02$ but degrades on fMRI. The superior performance of the graph-based generative models can be attributed to their explicit topological modeling using GATv2 encoders, which preserves the structural properties of sFNC matrices better than vectorized representations. The MoE fusion mechanism and latent space learning also enable robust multimodal integration compared to end-to-end diffusion approaches. Identical performance across DiT, LDM, and MMVAE indicates that the advantage of these models is due to the VAE-based latent representations rather than specific diffusion architectures. MMVAE also better preserves the variance patterns compared to DiT and LDM (Fig.~\ref{fig:overview}). 

\begin{table*}[ht]
\centering
\caption{Comparison of reconstruction evaluation performance across fMRI and sMRI modalities for various generative models}
\label{fig:recon}
\begin{adjustbox}{max width=\textwidth}
\begin{tabular}{|l|lllll|llll|}
\hline
\multirow{2}{*}{\textbf{Model}} & \multicolumn{5}{c|}{\textbf{fMRI}}                                                                                                                                                                                                                  & \multicolumn{4}{c|}{\textbf{sMRI}}                                                                                                                                                                     \\ \cline{2-10} 
                                & \multicolumn{1}{l|}{\textbf{MSE}}         & \multicolumn{1}{l|}{\textbf{\begin{tabular}[c]{@{}l@{}}Frobenius \\ Norm\end{tabular}}} & \multicolumn{1}{l|}{\textbf{Correlation}} & \multicolumn{1}{l|}{\textbf{SSIM}}        & \textbf{PSNR}         & \multicolumn{1}{l|}{\textbf{MSE}}         & \multicolumn{1}{l|}{\textbf{\begin{tabular}[c]{@{}l@{}}Frobenius\\ Norm\end{tabular}}} & \multicolumn{1}{l|}{\textbf{Correlation}} & \textbf{PSNR}         \\ \hline
\textbf{tDDPM}      & \multicolumn{1}{l|}{0.13+/-0.03}          & \multicolumn{1}{l|}{18.98+/-2.12}                                                       & \multicolumn{1}{l|}{0.33+/-0.09}          & \multicolumn{1}{l|}{0.11+/-0.06}          & 14.47+/-0.94          & \multicolumn{1}{l|}{63.11+/-13.67}        & \multicolumn{1}{l|}{57.50+/-6.19}                                                      & \multicolumn{1}{l|}{0.33+/-0.23}          & 6.09+/-0.66           \\ \hline
\textbf{WGAN-GP}                & \multicolumn{1}{l|}{0.11+/-0.03}          & \multicolumn{1}{l|}{17.17+/-2.61}                                                       & \multicolumn{1}{l|}{0.53+/-0.09}          & \multicolumn{1}{l|}{0.29+/-0.09}          & 16.19+/-1.23          & \multicolumn{1}{l|}{2.40+/-3.89}          & \multicolumn{1}{l|}{9.86+/-5.46}                                                       & \multicolumn{1}{l|}{0.97+/-0.02}          & 23.01+/-4.30          \\ \hline
\textbf{vDiT}                    & \multicolumn{1}{l|}{0.04+/-0.01} & \multicolumn{1}{l|}{10.84+/-1.43}                                                        & \multicolumn{1}{l|}{0.79+/-0.06}          & \multicolumn{1}{l|}{0.51+/-0.09}          & 18.61+/-1.16          & \multicolumn{1}{l|}{0.24+/-0.37}          & \multicolumn{1}{l|}{3.41+/-1.15}                                                       & \multicolumn{1}{l|}{\textbf{0.99+/-0.01}} & 31.07+/-1.96          \\ \hline
\textbf{vLDM}                    & \multicolumn{1}{l|}{0.04+/-0.01} & \multicolumn{1}{l|}{10.85+/-1.44}                                                        & \multicolumn{1}{l|}{0.79+/-0.06} & \multicolumn{1}{l|}{0.51+/-0.10} & 18.61+/-1.05          & \multicolumn{1}{l|}{0.24+/-0.36} & \multicolumn{1}{l|}{3.40+/-1.14}                                                       & \multicolumn{1}{l|}{\textbf{0.99+/-0.01}} & 31.09+/-1.95          \\ \hline
\textbf{vMMVAE}                  & \multicolumn{1}{l|}{0.04+/-0.01} & \multicolumn{1}{l|}{10.84+/-1.44}                                               & \multicolumn{1}{l|}{0.79+/-0.06} & \multicolumn{1}{l|}{0.51+/-0.10} & 18.62+/-1.05 & \multicolumn{1}{l|}{0.24+/-0.36} & \multicolumn{1}{l|}{3.38+/-1.13}                                              & \multicolumn{1}{l|}{\textbf{0.99+/-0.01}} & 31.15+/-1.92 \\ \hline
\textbf{gDiT}                    & \multicolumn{1}{l|}{\textbf{0.03+/-0.01}} & \multicolumn{1}{l|}{9.17+/-1.15}                                                        & \multicolumn{1}{l|}{0.85+/-0.05}          & \multicolumn{1}{l|}{0.63+/-0.09}          & 20.08+/-1.00          & \multicolumn{1}{l|}{0.31+/-1.42}          & \multicolumn{1}{l|}{3.58+/-1.89}                                                       & \multicolumn{1}{l|}{\textbf{0.99+/-0.01}} & 30.81+/-2.11          \\ \hline
\textbf{gLDM}                    & \multicolumn{1}{l|}{\textbf{0.03+/-0.01}} & \multicolumn{1}{l|}{8.60+/-1.08}                                                        & \multicolumn{1}{l|}{\textbf{0.87+/-0.04}} & \multicolumn{1}{l|}{\textbf{0.67+/-0.08}} & 20.67+/-0.99          & \multicolumn{1}{l|}{\textbf{0.23+/-0.33}} & \multicolumn{1}{l|}{3.33+/-1.12}                                                       & \multicolumn{1}{l|}{\textbf{0.99+/-0.01}} & 31.29+/-1.92          \\ \hline
\textbf{gMMVAE}                  & \multicolumn{1}{l|}{\textbf{0.03+/-0.01}} & \multicolumn{1}{l|}{\textbf{8.57+/-1.08}}                                               & \multicolumn{1}{l|}{\textbf{0.87+/-0.04}} & \multicolumn{1}{l|}{\textbf{0.67+/-0.08}} & \textbf{20.70+/-1.00} & \multicolumn{1}{l|}{\textbf{0.23+/-0.33}} & \multicolumn{1}{l|}{\textbf{3.29+/-1.12}}                                              & \multicolumn{1}{l|}{\textbf{0.99+/-0.01}} & \textbf{31.38+/-1.92} \\ \hline
\end{tabular}
\end{adjustbox}
\end{table*}
\vspace{-12pt}
\subsection{Computational efficiency analysis}
\vspace{-6pt}
A comprehensive efficiency analysis was conducted across the different generative models to evaluate the model complexity and computational cost. Table.~\ref{fig:eff} demonstrates that vMMVAE-based models achieve the lowest parameter count, floating-point operations (FLOPs), and latency amongst all the architectures. Similarly, the graph-based variant, gMMVAE, maintains high efficiency with only 9.22M parameters and 31ms latency, representing a reduction in latency compared to gLDM and gDiT. The diffusion-based models have substantially high computational overhead due to iterative denoising across multiple timesteps. As a result, despite having only 35.31M parameters, tDDPM remains computationally expensive for real-world applications due to high FLOPs and latency. Among the diffusion variants, LDM models achieve the lowest FLOPs by operating in a compressed latent space. Meanwhile, WGAN-GP emerges as the most lightweight baseline, with only 2.06M parameters and the lowest FLOPs. However, its poor reconstruction quality, as shown in Table.~\ref{fig:recon} indicates that this computational advantage does not translate into effective performance. In terms of inference time, the graph-based models show modestly higher inference times compared to the vectorized MLP variants due to the GATv2 operations. Overall, this efficiency analysis highlights that MMVAE’s single-pass generation provides faster sampling than diffusion-based alternatives, achieving comparable or superior reconstruction quality in most cases (Table.~\ref{fig:recon}).

\begin{table}[htbp]
\centering
\caption{Efficiency analysis across various generative models}
\label{fig:eff}
\begin{adjustbox}{max width=\columnwidth}
\begin{tabular}{|c|l|l|l|l|}
\hline
\textbf{Models}            & \multicolumn{1}{c|}{\textbf{\begin{tabular}[c]{@{}c@{}}Number of\\ parameters (M)\end{tabular}}} & \multicolumn{1}{c|}{\textbf{\begin{tabular}[c]{@{}c@{}}FLOPS\\ (G)\end{tabular}}} & \multicolumn{1}{c|}{\textbf{\begin{tabular}[c]{@{}c@{}}Latency\\  (ms)\end{tabular}}} & \multicolumn{1}{c|}{\textbf{\begin{tabular}[c]{@{}c@{}}Inference \\  time (ms)\end{tabular}}} \\ \hline
\textbf{tDDPM} & 35.31                                                                                            & 125                                                                              & 1860                                                                          & 0.22                                                                                          \\ \hline
\textbf{WGAN-GP}           & 2.06                                                                                             & 0.03                                                                            & 0.90                                                                           & 0.24                                                                                          \\ \hline
\textbf{vDiT}               & 1.85                                                                                            & 3.95                                                                            & 542                                                                           & 0.31                                                                                          \\ \hline
\textbf{vLDM}               & 0.8                                                                                             & 0.15                                                                            & 107                                                                           & 0.30                                                                                          \\ \hline
\textbf{vMMVAE}             & 0.71                                                                                             & 0.01                                                                           & 5                                                                         & 0.08                                                                                         \\ \hline
\textbf{gDiT}               & 10.36                                                                                            & 3.95                                                                            & 544                                                                           & 0.32                                                                                          \\ \hline
\textbf{gLDM}               & 9.31                                                                                             & 0.14                                                                            & 146                                                                           & 0.41                                                                                          \\ \hline
\textbf{gMMVAE}             & 9.22                                                                                             & 0.02                                                                            & 31                                                                         & 0.14                                                                                         \\ \hline
\end{tabular}
\end{adjustbox}
\end{table}
\vspace{-12pt}
\subsection{Generation quality assessment}
\vspace{-6pt}
The generative quality of the three best-performing graph-based models is summarized in Table.~\ref{fig:gen}. Here, the various metrics capture complementary aspects of generative quality, with greater emphasis on WD and KL divergence for global alignment and MMD for local similarity. Each model generated 10000 samples, which were compared against the UK Biobank test data. For fMRI, gDiT achieved the best MMD score, demonstrating 9.7\% improvement over gLDM and 34\% over gMMVAE. However, gMMVAE outperformed diffusion variants in WD and KL divergence and better preserved the topological properties of brain connectivity. For sMRI generation, gMMVAE demonstrated superiority with lower MMD values and KL divergence. gMMVAE achieved a 42\% improvement in KL over gDiT and 22\% over gLDM, although gDiT attained a better WD. The superior MMD performance of gDiT for fMRI indicates that the transformer-based attention mechanism better captures local correlations in the latent space. While the advantage of gMMVAE in WD and KL metrics indicates that it accurately models the overall distribution structure of connectivity data. To further validate these distribution characteristics, the 10,000 samples generated by each model were compared against the true test data after computing the mean and standard deviation of the overall data. The visual analysis from Fig.~\ref{fig:overview} confirms that gMMVAE produces samples with distribution statistics most closely aligned with the real data for both modalities, while gDiT and gLDM show marginal but uniform overestimation of variance in the connectivity patterns. Overall, gMMVAE provides the best generation quality through direct sampling. 

\vspace{-12pt}
\begin{table}[ht]
\centering
\caption{Generation quality analysis across models}
\label{fig:gen}
\begin{adjustbox}{max width=\columnwidth}
\begin{tabular}{|l|l|l|l|l|}
\hline
\textbf{Modality}              & \textbf{Metric} & \textbf{DiT}    & \textbf{LDM} & \textbf{MMVAE}  \\ \hline
\multirow{3}{*}{\textbf{fMRI}} & MMD             & \textbf{0.0215} & 0.0238       & 0.0326          \\ \cline{2-5} 
                               & WD              & 0.1851          & 0.1866       & \textbf{0.1259} \\ \cline{2-5} 
                               & KL              & 8.0526          & 8.0002       & \textbf{4.0045}  \\ \hline
\multirow{3}{*}{\textbf{sMRI}} & MMD             & 1.2751          & 1.2761       & \textbf{0.7848} \\ \cline{2-5} 
                               & WD              & \textbf{2.7466} & 2.7920       & 3.1965          \\ \cline{2-5} 
                               & KL              & 0.9201          & 0.6810       & \textbf{0.5296} \\ \hline
\end{tabular}
\end{adjustbox}
\end{table}
\vspace{-20pt}
\subsection{Ablation study}
\vspace{-6pt}
The analysis from Table.~\ref{fig:lat} shows that incorporating conditioning information into both encoder and decoder stages consistently improves classification performance across all models. In the case of gMMVAE, conditioning from the encoder only to dual conditioning yields modest improvements, with accuracy increasing by 1.1\% and F1 score improving by 1.2\%. Meanwhile, gLDM shows the largest improvement from dual conditioning, achieving 6.2\% higher accuracy and a 6.1\% increase in F1 score compared to encoder-only conditioning. This dual-conditioned gLDM model also achieves the highest overall precision at 92.44\%. Conversely, gDiT exhibits intermediate improvements, with dual conditioning providing a 3\% accuracy gain and a 2.8\% increase in F1 score, along with the highest overall precision score of 92.55\%. In the case of dual conditioning, gDiT achieves the best overall accuracy, marginally outperforming the other models. In contrast, encoder-only conditioning reveals significant performance differences, with gLDM showing the lowest accuracy at 84.65\%, compared to gMMVAE’s 88.16\% and gDiT’s 87.87\%. These results indicate that diffusion-based models are dependent on decoder-level conditioning to learn discriminative latent representations. The substantial improvement with dual conditioning demonstrates that, during encoding, the conditioning guides the network to extract sex specific features from the multimodal neuroimaging data. To further probe this effect, we performed a gradient-based sensitivity analysis of the latent space, which indicates that these differences are distributed across functional connectivity patterns, with relatively higher contributions from default mode and cognitive control networks, consistent with prior findings of distributed sex-related effects.
\vspace{-12pt}
\begin{table}[htbp]
\centering
\caption{Latent space ablation study}
\label{fig:lat}
\begin{adjustbox}{max width=\columnwidth}
\begin{tabular}{|l|l|l|l|l|l|}
\hline
\textbf{Model}         & \textbf{Conditioning}                                          & \textbf{Accuracy} & \textbf{Precision} & \textbf{Recall} & \textbf{F1 Score} \\ \hline
\multirow{2}{*}{MMVAE} & Encoder only                                                   & 88.16+/-1.94      & 88.69+/-2.11       & 86.46+/-2.55    & 87.55+/-2.04      \\ \cline{2-6} 
                       & \begin{tabular}[c]{@{}l@{}}Encoder and \\ Decoder\end{tabular} & 89.17+/-1.38      & 90.14+/-2.26       & 87.08+/-1.56    & 88.57+/-1.37      \\ \hline
\multirow{2}{*}{LDM}   & Encoder only                                                   & 84.65+/-1.36      & 84.29+/-2.64       & 83.96+/-3.88    & 84.02+/-1.64      \\ \cline{2-6} 
                       & \begin{tabular}[c]{@{}l@{}}Encoder and \\ Decoder\end{tabular} & 89.87+/-0.98      & 92.44+/-1.98       & 86.04+/-1.25    & 89.11+/-0.97      \\ \hline
\multirow{2}{*}{DiT}   & Encoder only                                                   & 87.87+/-1.90      & 87.71+/-2.38       & 87.08+/-3.58    & 87.34+/-2.09      \\ \cline{2-6} 
                       & \begin{tabular}[c]{@{}l@{}}Encoder and \\ Decoder\end{tabular} & 90.47+/-1.33      & 92.55+/-1.85       & 87.29+/-2.90    & 89.80+/-1.54      \\ \hline
\end{tabular}
\end{adjustbox}
\end{table}

\vspace{-20pt}
\section{Conclusion}
\label{sec:conclude}
\vspace{-6pt}
This paper presents graph-based generative frameworks for multimodal neuroimaging fusion. Comprehensive evaluation on a large dataset demonstrates that graph-based generative models substantially outperform vectorized approaches. The proposed gMMVAE efficiently reconstructs and generates both modalities with high fidelity, surpassing baseline variants. The GATv2 encoder effectively preserves brain network topology, while dual conditioning is crucial for learning discriminative latent representations, enhancing classification accuracy, particularly for diffusion models. The mixture-of-experts fusion mechanism effectively integrates heterogeneous modalities into a unified latent space, enabling efficient multimodal fusion. Future directions include extending to voxel-level representations through scalable graph constructions, incorporating additional modalities, finer-grained phenotypic conditioning, and validation on disease-specific datasets. While the proposed gMMVAE offers optimal efficiency for multimodal integration, it also serves as a scalable foundation with promising extensions to longitudinal studies, disease modeling, and large-scale population analyses.

\vspace{-8pt}
\section{Acknowledgments}
\vspace{-12pt}
This work was supported by NIH grants 1R01AG090597, R01AG073949 awarded to Vince Calhoun.
\vspace{-12pt}
\bibliographystyle{IEEEbib}
\bibliography{refs}

\end{document}